\pdfoutput=1
\documentclass[10pt,twocolumn,letterpaper]{article}

\usepackage[pagenumbers]{cvpr}

\usepackage{tabularray}
\UseTblrLibrary{booktabs}
\usepackage{tikz}
\usepackage{pgfplots}
\pgfplotsset{compat=1.17}
\usetikzlibrary{patterns,arrows.meta,positioning,calc}
\usepackage{algorithm}
\usepackage{algorithmic}

\newcommand{\gc}{ViewCache}

\newcommand{\xz}{x_0}

\definecolor{cvprblue}{rgb}{0.21,0.49,0.74}
\usepackage[pagebackref,breaklinks,colorlinks,allcolors=cvprblue]{hyperref}

\title{ViewCache: Training-Free Acceleration of Multi-View Texture Diffusion via Adaptive View Selection}

\author{
Haotang Li\textsuperscript{1} \quad
Zhenyu Qi\textsuperscript{1} \quad
Shaohan Henry Wang\textsuperscript{1} \quad
Kebin Peng\textsuperscript{2} \quad
Yutong Zhao\textsuperscript{3}\\
Zi Wang\textsuperscript{4} \quad
Bo Liu\textsuperscript{1} \quad
Huanrui Yang\textsuperscript{1} \quad
Sen He\textsuperscript{1}\\[4pt]
{\normalsize \textsuperscript{1}Department of Electrical and Computer Engineering, University of Arizona, Tucson, AZ}\\
{\normalsize \textsuperscript{2}Department of Computer Science, East Carolina University, Greenville, NC}\\
{\normalsize \textsuperscript{3}Department of Computer Engineering and Computer Science,California State University, Long Beach, CA}\\
{\normalsize \textsuperscript{4}School of Computer and Cyber Sciences, Augusta University, Augusta, GA}\\
}

\begin{document}
\maketitle
\begin{abstract}
Multi-view diffusion produces high-quality textures for 3D assets, but every denoising step evaluates the network on all views, which makes generation slow.
Existing training-free accelerators treat the views of a step as a whole, so a step is skipped or approximated for all views together.
We observe that the view level offers a finer unit.
Within a step, the views differ in how much they change, and views that show the same surface are similar there.
Based on these observations, we propose \gc{}, a training-free accelerator that scores the importance of each view and selects any number of views to evaluate at each step, from none to all, under a synchronization constraint. The remaining views are restored without evaluating the network.
On four multi-view texturing backbones, \gc{} achieves a better trade-off between speedup and fidelity than six state-of-the-art training-free accelerators.
Within the same fidelity budget, it reaches \(2.72\times\) against \(2.21\times\) on Hunyuan3D-2.1 and \(4.47\times\) against \(3.65\times\) on MV-Adapter.
It can also be combined with step-level accelerators, which makes them both faster and closer to the original model.
\end{abstract}

\suppressfloats[t]
\begin{figure}[t]
\centering
\input{Figures/f4_ladder_data}%
\resizebox{\columnwidth}{!}{\input{Figures/f1_teaser_tikz}}
\caption{(a) Step-level SOTAs evaluate all \(N\) views of a step or none, while \gc{} evaluates any number. (b) Trade-off on Hunyuan3D-2.1, with a view, a zoom, and the error to stock.}
\label{fig:teaser}
\end{figure}

\section{Introduction}
\label{sec:intro}

Text- and image-conditioned 3D asset generation has converged on a two-stage recipe, in which a shape model produces geometry \citep{clay,trellis} and a geometry-conditioned multi-view diffusion model paints it, with the painted views baked into a UV texture \citep{hunyuan3d21,mvadapter,mvpainter}.
Painting takes a large share of the production cost, 47.5\% of the end-to-end wall-clock time on Hunyuan3D-2.1 and 56.2\% on SyncMVD at the median asset and at production resolution, and within it the denoising loop is the component that a training-free accelerator can act on.
This loop evaluates the denoiser on \(N\) views at each of \(T\) steps.
Its schedule therefore has two levels.
The step level decides which steps are computed, and the view level decides which views are computed within each step.

Training-free accelerators reduce the cost of the \(N \times T\) view evaluations of the denoising loop without retraining the model.
Some skip denoising steps, computing only a subset of the \(T\) steps and reusing or forecasting the outputs of the skipped steps, such as TeaCache, MagCache, FORA, TaylorSeer, and BudCache \citep{teacache,magcache,fora,taylorseer,budcache}.
Some keep all \(T\) steps and make each of them cheaper by reusing part of the network within the step, such as FasterCache and DeepCache \citep{fastercache,deepcache}.
They reach substantial speedups with little loss of quality on image and video generation.
However, all of them treat the \(N\) views of a step as a whole.
In these approaches, one decision applies to all \(N\) views at each of the \(T\) steps. A step is either kept or skipped for all views (\cref{fig:teaser}a, with illustrative schedules), or the same part of the network is reused for all views, so the approximation degrades all \(N\) views together.
Current methods show that the step level can be optimized effectively, but \textbf{a finer-grained optimization at the view level remains under-explored.}
We conduct a motivating study (\cref{sec:preliminary}) to explore acceleration at the view level and make two key observations.
First, the \(N\) views do not change equally at each step.
Between consecutive steps, some views remain almost static while others change substantially, and the set of rapidly changing views shifts across the \(T\) steps.
Evaluating either all \(N\) views or none is therefore too coarse, and we should instead select views individually at each step.
Second, overlapping views share similar updates on the object surface.
Since all views depict the same 3D object, most surface points are visible in multiple views and receive similar predictions and step updates.
The views therefore cannot be selected independently, and the unevaluated views can be restored from the evaluated ones.
Our motivating study shows that a schedule that evaluates only the views whose change exceeds a threshold at each step achieves a lower error than step-level accelerators at the same speedup (\cref{fig:redundancy}d).

Based on these observations, we propose \gc{}, a training-free accelerator that operates at the view level.
At each denoising step, \gc{} scores the importance of every view and selects any number of views to evaluate, from none to all \(N\).
A synchronization constraint in the selection keeps views that overlap on the surface consistent with each other.
\gc{} then restores the views that are not evaluated without calling the network (\cref{fig:teaser}a).
\Cref{fig:method} gives an overview of the method.
This schedule is finer-grained than that of step-level methods, which evaluate either all views or none at each step.
Since the view-level decision is made within a step, it is independent of how steps are selected.
\gc{} is therefore general, and it can also be combined with a step-level accelerator without changing the decision rule of that accelerator (\cref{sec:generality}).

We evaluate \gc{} on four multi-view texturing backbones, namely Hunyuan3D-2.1, MV-Adapter, SyncMVD, and MVPainter, against six training-free accelerators.
Within a fidelity budget of 0.03 in MV-LPIPS, \gc{} reaches a higher speedup than the best of these accelerators on all four backbones, for example \(2.72\times\) against \(2.21\times\) on Hunyuan3D-2.1 and \(4.47\times\) against \(3.65\times\) on MV-Adapter (\cref{fig:speedperf}).
Combined with a step-level accelerator, \gc{} makes it both faster and closer to the original model.
With TeaCache on Hunyuan3D-2.1, the speedup rises from \(2.00\times\) to \(2.30\times\) while the MV-LPIPS drops from 0.0261 to 0.0156 (\cref{fig:analysis}b).

Our contributions are summarized as follows.
1) We identify the view level as a finer unit of training-free acceleration for multi-view diffusion.
2) We propose \gc{}, a view-level training-free accelerator that selects the views of each step by an importance score under a synchronization constraint and restores the others without network evaluations.
3) Evaluations on four backbones show that \gc{} achieves a better trade-off between speedup and fidelity than state-of-the-art training-free accelerators, and that it can be combined with step-level accelerators.

\section{Related Work}
\label{sec:related}

\noindent\textbf{Training-free diffusion acceleration.}
Most training-free diffusion caches act at the step level and decide which denoising steps run the full network.
They differ in the reuse trigger, which is timestep-embedding drift \citep{teacache}, residual-magnitude ratios \citep{magcache}, a fixed interval \citep{fora}, a Taylor forecast of skipped features \citep{taylorseer}, or differences between feature maps \citep{deltadit}.
BudCache fixes the number of computed steps in advance and searches offline for the skip schedule that best preserves the output \citep{budcache}.
Fewer-step solvers \citep{unipc,ddim,dpmsolver} and step distillation \citep{progressive,consistency,lcm,dmd,add} shorten the trajectory instead of skipping within it.
A second family acts within a step and reduces the cost of each computed step. DeepCache reuses deep U-Net features \citep{deepcache}, FasterCache reuses the classifier-free-guidance branch \citep{fastercache}, and ToCa selects which tokens to recompute \citep{toca}.
The survey of \citet{cachesurvey} organizes diffusion caching into temporal, architectural, and token-level reuse.
On a multi-view backbone, both families evaluate all \(N\) views at every computed step.
\gc{} instead selects which views are evaluated at each step, a decision that these methods do not make and with which they can be combined.

\noindent\textbf{Non-uniform schedules across a batch.}
RhymeFlow accelerates video diffusion by giving frames different schedules, in which a sparse set of keyframes is denoised densely and the other frames skip steps and are recovered by latent trajectory projection \citep{rhymeflow}.
Its results show that the units of a jointly denoised batch need not share one schedule, and that such a per-unit schedule composes with cache-style accelerators.
RhymeFlow relates frames through temporal adjacency in a video model.
Hash3D reuses features between nearby camera poses during score-distillation \citep{dreamfusion} 3D generation \citep{hash3d}, and reverse reprojection caching in rendering transports a per-pixel quantity through geometric correspondence and refreshes it periodically \citep{reproj}.
Unlike these methods, \gc{} gives the views of a 3D asset different schedules inside a denoising trajectory, selects the views of every step by an importance score, and keeps overlapping views synchronized through position maps.

\begin{figure*}[t]
    \centering
    \resizebox{\textwidth}{!}{\input{Figures/f2_motivation_tikz}}
    \caption{Motivating study on Hunyuan3D-2.1. (a) A step-level accelerator pushed to \({\sim}2.2\times\) changes all six views. (b) The change of the predicted clean signal \(\xz\) of every view at every step, normalized at each step, where darker is larger. (c) The similarity between views at corresponding points over the steps. (d) MV-LPIPS of step-level accelerators and of a view-level schedule at a speedup of about \(2\times\).}
    \label{fig:redundancy}
\end{figure*}

\noindent\textbf{Multi-view texture generation.}
Hunyuan3D-2.1 Paint \citep{hunyuan3d21,hunyuan3d2} and MV-Adapter \citep{mvadapter} denoise per-view latents in one batch with multi-view attention, following multi-view diffusion models for 3D generation \citep{mvdream,syncdreamer}. MVPainter \citep{mvpainter} arranges the views as the tiles of one grid image, and SeqTex \citep{seqtex} treats views as video frames denoised jointly with a UV map.
MVDiffusion adds correspondence-aware attention between geometrically related views \citep{mvdiffusion}, and CAMEO and CaliTex use geometric structure to supervise or calibrate multi-view attention \citep{cameo,calitex}.
SyncMVD \citep{syncmvd} and MD-ProjTex \citep{mdprojtex} synchronize views through a shared UV representation.
TEXGen \citep{texgen} denoises a single UV map, and optimization- and inpainting-based methods \citep{texture,text2tex,texfusion,paint3d} generate views sequentially.
These methods design how the views are generated and kept consistent, and they evaluate all \(N\) views at every step.
\gc{} leaves the model unchanged and evaluates only the selected views, and it updates each remaining view from its own previous state and the update transported from the evaluated views.

\section{Motivating Study}
\label{sec:preliminary}
\noindent\textbf{Optimization on multi-view diffusion.}
Given a mesh, a geometry-conditioned multi-view diffusion model jointly denoises \(N\) view latents \(\{x_t^{(v)}\}_{v=1}^{N}\) over \(T\) sampler steps, and the denoised views are baked into a UV texture \citep{hunyuan3d21,mvadapter,mvpainter}.
At each step, the denoiser processes the \(N\) views as one batch, in which the views interact, typically through multi-view attention.
It is conditioned on rendered geometry maps, including canonical coordinate (position) maps, which provide the 3D coordinate observed by each latent token and therefore associate tokens in different views that observe the same surface point.
We write \(\xz^{(v)}(t)\) for the clean signal of view \(v\) predicted at step \(t\), which is obtained in closed form from the noisy latent and the network output.
\(\mathcal{C}\) denotes the set of steps at which the denoiser runs, \(\mathcal{V}_t\) denotes the set of views it runs on at step \(t\), and \(c(n)\) denotes the cost of one denoiser call on \(n\) views, so the loop cost is
\begin{equation}
\mathrm{Cost} \;=\; \sum\nolimits_{t\in\mathcal{C}} c\big(|\mathcal{V}_t|\big) .
\label{eq:cost}
\end{equation}
\emph{Step-level} methods prune steps and reuse or forecast their outputs, which reduces \(|\mathcal{C}|\) \citep{teacache,magcache,budcache,fora,taylorseer}.
\emph{In-step} methods reuse part of the network within a step, which reduces \(c(\cdot)\) \citep{deepcache,fastercache}.
Both keep \(\mathcal{V}_t\) equal to all \(N\) views.
\emph{View-level} acceleration, which we study in this paper, reduces \(|\mathcal{V}_t|\), where \(\mathcal{V}_t\) can be any subset of the \(N\) views, from none to all.
Step-level methods are the special case where \(\mathcal{V}_t\) is empty or full, so a view-level schedule is finer-grained.

\noindent\textbf{Observations.}
Skipping a step degrades all \(N\) views simultaneously because a step-level accelerator makes a single decision for the entire batch at each step.
On the asset in \cref{fig:redundancy}a, pushing MagCache to a \(2.21\times\) speedup raises the MV-LPIPS of all six views to between 0.050 and 0.060.
We therefore explore whether a step can evaluate only a subset of its \(N\) views.
This approach requires that some views need evaluation less than others at a given step, and that the views are closely related so that unevaluated views can still be updated.
We investigate both premises on dev20 using Hunyuan3D-2.1 and the metrics in \cref{sec:setup}, while the supplement provides the full protocol.

\emph{First, views vary in how much they change within the same step.}
We track the change in \(\xz^{(v)}\) between consecutive steps for each view during a full evaluation run (\cref{fig:redundancy}b).
Within a single step, the largest change among the \(N\) views is \(4.47\) times the smallest on average, with a median of \(2.55\) times. The figure averages the changes over the assets, so its contrast is lower than that of a single asset.
View importance also shifts dynamically across steps because the view with the largest change differs from that of the previous step in 30.8\% of the steps, which rules out a static schedule for each view.
This ranking can be predicted beforehand because the change of a view at the previous step achieves a rank correlation of 0.763 with its change at the current step.
Evaluating either all \(N\) views or none is therefore too coarse, and the views should be selected adaptively.

\begin{figure*}[t]
    \centering
    \resizebox{\textwidth}{!}{\input{Figures/f3_method_tikz}}
    \caption{Overview of \gc{}. (a) \gc{} acts inside the denoising loop and leaves the denoiser, the sampler, and the baking unchanged. (b) The selection scores every view by its predicted error and trades the error of the views left out against the cost. (c) The synchronization constraint penalizes overlapping views left with different errors. (d) A view that is not evaluated is updated by transport.}
    \label{fig:method}
\end{figure*}

\emph{Second, views exhibit strong similarity across shared surface regions.}
Position maps link pixels across different views that observe the same 3D surface point because all views render the same underlying mesh.
An average of 87.5\% of the foreground pixels in a view are visible in at least one other view.
At these corresponding tokens, we compute the cosine similarity between a view and the other views for both \(\xz\) and its step change at each step, using a shuffled correspondence as a control (\cref{fig:redundancy}c).
The \(\xz\) of a view remains consistent with other views at corresponding tokens throughout the trajectory, maintaining a cosine similarity of 0.83 to 0.87 from the third step onward compared to 0.79 to 0.82 for the control.
The step change of \(\xz\) also shows higher similarity at corresponding tokens than the control from the third step onward, gradually declining from 0.71 at the third step to 0.27 at the eighth step and 0.08 at the final step, compared to 0.64, 0.11, and 0.04 for the control.
Views are therefore correlated at corresponding surface points throughout denoising, so the selection must keep overlapping views consistent, and unevaluated views can be updated from evaluated ones.

These findings indicate that not every view needs to be evaluated at every step.
As a proof of concept, a view-level schedule that evaluates only the views whose measured change exceeds a threshold at each step achieves an MV-LPIPS of 0.0175 at a \(1.96\times\) speedup, outperforming TeaCache, MagCache, and BudCache under similar speedups (\cref{fig:redundancy}d).
The remaining challenges are how to select views without this prior knowledge and how to maintain consistency across views, and \gc{} addresses both.

\section{ViewCache}
\label{sec:method}
Our motivating study highlights two key tasks for accelerating diffusion at the view level.
The method must first decide which views to evaluate at each step because view importance shifts across steps.
The method must also update the unevaluated views because the sampler requires predictions for all \(N\) views.
\gc{} addresses these two tasks through adaptive view selection and a restoration of the unevaluated views.
\Cref{fig:method} presents an overview of our pipeline.
Inside the denoising loop, \gc{} inserts a selection stage before the denoiser and a transport stage after it, while leaving the denoiser, the sampler, and the texture baking unchanged (\cref{fig:method}a).
\gc{} maintains the predicted clean signal \(\xz^{(v)}\) on the latent token grid for each guidance and material branch of every view.

At each denoising step \(t\), \gc{} operates in three stages.
First, it selects a subset of views \(\mathcal{V}_t\) to evaluate, ranging from zero to all \(N\) views.
This selection relies on an importance score tracked during the current generation (\cref{sec:select}) and a synchronization constraint that preserves consistency across overlapping views (\cref{sec:sync}).
Second, \gc{} runs the denoiser only on the selected subset \(\mathcal{V}_t\) by slicing the input batch.
During this forward pass, attention spans the \(|\mathcal{V}_t|\) active views while each view retains its original positional indices.
Third, \gc{} updates the predictions of the remaining unevaluated views via geometric transport without additional network evaluations (\cref{sec:transport}).
After these three stages, the sampler receives the complete set of \(N\) view predictions and updates the latents as usual.
The entire process requires no training and modifies no model weights.

\subsection{Importance-Scored View Selection}
\label{sec:select}
The views of a step are not equally important, and the important views change from step to step (\cref{sec:preliminary}).
\gc{} therefore decides at every step which views are evaluated and how many.
This decision is made before the denoiser runs, so it can only use quantities that are already available.

\noindent\textbf{Importance score.}
For view \(v\), \(\tau_v\) denotes the last step at which it was evaluated, \(g_v\) denotes the number of steps between its last two evaluations, and \(E_v\) denotes the mean squared difference of \(\xz^{(v)}\) between these two evaluations over the foreground tokens, which are the tokens that show the object.
We assume that \(\xz^{(v)}\) keeps drifting linearly at its most recent rate.
The error of keeping the last prediction of view \(v\) at step \(t\) is then
\begin{equation}
e_v(t) \;=\; E_v \left( \frac{t - \tau_v}{g_v} \right)^{2}.
\label{eq:score}
\end{equation}
The predicted error \(e_v(t)\) is used as the importance of view \(v\) at step \(t\).
In the example of \cref{fig:method}b, views \(v_2\), \(v_3\), and \(v_6\) have the largest predicted errors.
It grows quadratically with the number of steps since the last evaluation, so every view is selected again after some steps and no periodic refresh is needed.
The score is not defined before a view has been evaluated twice, so the first two steps evaluate all \(N\) views, which are the first two columns of the schedule in \cref{fig:teaser}a.

\noindent\textbf{Selection.}
The cost \(c(n)\) of \cref{eq:cost} is measured as the time of one denoiser step on \(n\) views, with \(c(0)=0\), and it need not be proportional to \(n\).
Evaluating a set \(\mathcal{V}\) of views leaves the predicted error of the other views and costs \(c(|\mathcal{V}|)\), and \gc{} trades the two with one weight \(\lambda\),
\begin{equation}
J_0(\mathcal{V}) \;=\; \sum\nolimits_{v \notin \mathcal{V}} e_v(t) \;+\; \lambda\,\frac{c(|\mathcal{V}|)}{c(N)} .
\label{eq:select}
\end{equation}
The cost depends only on how many views are evaluated, and the error is the sum of the scores of the views that are not evaluated.
For a given number \(n\) of evaluated views, the error is therefore smallest when the \(n\) views with the highest scores are the ones that are evaluated.
Minimizing \(J_0\) then only requires comparing the \(N+1\) possible values of \(n\), from \(0\) to \(N\).
The curves in \cref{fig:method}b show this comparison.
Evaluating more views leaves less error and takes more time.
The sum of the two is smallest at \(n=3\), which the figure marks as \(n^\star\), so \(v_2\), \(v_3\), and \(v_6\) are evaluated and the other three views are not.
The empty set skips the step for all views, and the full set is a step of the stock model.
\(\lambda\) is the only speed parameter. With \(\lambda=0\), every view is evaluated at every step and \gc{} reproduces the stock sampler, and a larger \(\lambda\) evaluates fewer views.

\subsection{Synchronization Constraint}
\label{sec:sync}
The score of \cref{eq:score} treats every view on its own, but the views overlap on the surface.
When one of two overlapping views is evaluated at a step and the other is not, the two views no longer agree on the surface region that both of them show.
After baking, this disagreement can appear as a seam.
The selection therefore also has to account for which views are evaluated together, and we add the predicted disagreement to the selection objective,
\begin{equation}
J(\mathcal{V}) \;=\; J_0(\mathcal{V}) \;+\; \mu \sum\nolimits_{u<v} O_{uv}\,\big(r_u(\mathcal{V}) - r_v(\mathcal{V})\big)^{2} .
\label{eq:sync}
\end{equation}
Here \(\mu\) is the weight of the constraint, and \(r_v(\mathcal{V})\) is the error magnitude that view \(v\) carries after the step, which is \(0\) if \(v \in \mathcal{V}\) and \(\sqrt{e_v(t)}\) otherwise.
\(O_{uv}\) is the overlap of views \(u\) and \(v\), which is the share of the foreground tokens of one view that show a surface point also shown in the other view, averaged over the two directions.
It is computed once per asset from the correspondence between views that \cref{sec:transport} describes.
The new term is large when one of two strongly overlapping views is evaluated while the other is left with a large predicted error, and it vanishes when the two views carry the same error.
In \cref{fig:method}c, the links show the overlap between the six views of the example, and the thickest link is the penalized pair, in which \(v_3\) is evaluated while \(v_4\), which overlaps with it strongly, is left out.
The objective now depends on which views are evaluated and not only on how many.
We evaluate \(J\) for all \(2^N\) subsets, which is cheap next to a denoiser call, and \(\mu=0\) recovers \cref{eq:select}.

\subsection{Restoring Unevaluated Views}
\label{sec:transport}
The sampler needs a prediction for all \(N\) views at every step, including the views that are not evaluated.
This step only completes the predictions and calls no network, so it is not a source of the speedup.
The motivating study shows that the views are similar at the corresponding points of the surface that they share, in their predictions and, in part, in the update of a step (\cref{sec:preliminary}).
The update of the evaluated views can therefore be transferred to a view that is not evaluated.
\gc{} restores such a view from its own previous prediction and from the update of the evaluated views.
For a view \(v\) outside \(\mathcal{V}_t\), we set
\begin{equation}
\xz^{(v)}(t) \;=\; \xz^{(v)}(t{-}1) \;+\; \omega\,\mathcal{G}_{\mathcal{V}_t\to v}\big[\,u_t\,\big],
\label{eq:delta}
\end{equation}
where \(u_t = \xz^{(\mathcal{V}_t)}(t) - \xz^{(\mathcal{V}_t)}(t{-}1)\) is the change of the evaluated views at this step and \(\omega\) is the transport weight.
The first term is a temporal transport, which carries the previous prediction of the view itself to the current step.
The second term is a geometric transport, which carries the change of the evaluated views at this step to view \(v\) through the surface correspondence \(\mathcal{G}_{\mathcal{V}_t\to v}\).
\Cref{fig:method}d illustrates it for a surface point \(p\) that is seen both in an evaluated view and in a view that is not evaluated. The change of the evaluated view at the token that shows \(p\) is carried to the token of the other view that shows \(p\), and is added to the previous prediction of that view.
When \(\mathcal{V}_t\) is empty, only the first term remains and the network is not called at all.

\begin{figure*}[t]
    \centering
    \input{Figures/f4_ladder_data}%
    \resizebox{\textwidth}{!}{\begin{tikzpicture}
\definecolor{gcred}{RGB}{200,50,45}
\definecolor{c1}{RGB}{54,125,189}\definecolor{c2}{RGB}{77,160,95}\definecolor{c3}{RGB}{140,100,180}
\definecolor{c4}{RGB}{214,120,40}\definecolor{c5}{RGB}{120,120,120}\definecolor{c6}{RGB}{170,140,60}
\pgfplotsset{
  lad/.style={width=5.2cm, height=4.05cm, enlarge x limits=0.04, grid=major, grid style={black!10},
    tick label style={font=\scriptsize}, label style={font=\footnotesize},
    ylabel style={at={(axis description cs:-0.20,0.5)}, anchor=south},
    enlarge y limits=0.08, scaled y ticks=false,
    yticklabel style={/pgf/number format/fixed}},
  ladlog/.style={log ticks with fixed point, ytick={0.001,0.002,0.005,0.01,0.02,0.05,0.1,0.2,0.5,1}},
  ladlogx/.style={xtick={1.5,2,3,5,8,13}, minor xtick={}},
  nox/.style={xticklabels={}},
  sTea/.style={c1, mark=o, mark size=1.8pt, line width=0.7pt},
  sMag/.style={c2, mark=square, mark size=1.7pt, line width=0.7pt},
  sBud/.style={c3, mark=triangle, mark size=2.1pt, line width=0.7pt},
  sFORA/.style={c4, mark=diamond, mark size=2.1pt, line width=0.7pt},
  sFaster/.style={c5, mark=pentagon, mark size=2pt, line width=0.7pt},
  sDeep/.style={c6, mark=x, mark size=2.2pt, line width=0.7pt},
  sGC/.style={gcred, mark=star, mark size=3pt, line width=1.3pt}}
\newcommand{\ladplot}[3]{\expandafter\ladploti\csname lad#2\endcsname{#1}{#3}}
\newcommand{\ladploti}[3]{\addplot[#2] table[x=speed,y=#3]{#1};}
\newcommand{\ladpanel}[2]{%
  \ladplot{sTea}{#1Tea}{#2}\ladplot{sMag}{#1Mag}{#2}\ladplot{sBud}{#1Bud}{#2}\ladplot{sFORA}{#1FORA}{#2}%
  \ladplot{sFaster}{#1Faster}{#2}\ladplot{sDeep}{#1Deep}{#2}\ladplot{sGC}{#1GC}{#2}}
\newcommand{\betterLR}{\draw[-{Stealth[length=3.5pt]},black!55,line width=0.6pt] (axis description cs:0.80,0.20) -- (axis description cs:0.94,0.06);
  \node[font=\scriptsize,text=black!55,anchor=south east,inner sep=1pt,fill=white] at (axis description cs:0.83,0.18) {better};}
\newcommand{\betterUR}{\draw[-{Stealth[length=3.5pt]},black!55,line width=0.6pt] (axis description cs:0.80,0.80) -- (axis description cs:0.94,0.94);
  \node[font=\scriptsize,text=black!55,anchor=north east,inner sep=1pt] at (axis description cs:0.83,0.82) {better};}
\newcommand{\budgetline}{\draw[black!60,densely dashed,line width=0.6pt] ({rel axis cs:0,0}|-{axis cs:2,0.03}) -- ({rel axis cs:1,0}|-{axis cs:2,0.03});}
\def\dx{0.88cm}\def\dy{-0.18cm}
\begin{axis}[ymode=log, lad, ladlog, nox, name=a1, ylabel={MV-LPIPS \(\downarrow\)},
  legend to name=ladderlegend, legend columns=7,
  legend style={font=\scriptsize, draw=black!30, /tikz/every even column/.append style={column sep=6pt}}]
\ladpanel{hy}{lpips}\budgetline\betterLR
\legend{TeaCache,MagCache,BudCache,FORA,FasterCache,DeepCache,\textbf{ViewCache}}
\end{axis}
\begin{axis}[ymode=log, lad, ladlog, nox, name=b1, at={(a1.east)}, anchor=west, xshift=\dx]\ladpanel{mva}{lpips}\budgetline\betterLR\end{axis}
\begin{axis}[ymode=log, lad, ladlog, nox, name=c1, at={(b1.east)}, anchor=west, xshift=\dx]\ladpanel{sync}{lpips}\budgetline\betterLR\end{axis}
\begin{axis}[ymode=log, xmode=log, lad, ladlog, ladlogx, nox, name=d1, at={(c1.east)}, anchor=west, xshift=\dx]\ladpanel{mvp}{lpips}\budgetline\betterLR\end{axis}
\begin{axis}[lad, name=a2, at={(a1.south)}, anchor=north, yshift=\dy, xlabel={Speedup (\(\times\))}, ylabel={MV-PSNR \(\uparrow\)}]\ladpanel{hy}{psnr}\betterUR\end{axis}
\begin{axis}[lad, name=b2, at={(b1.south)}, anchor=north, yshift=\dy, xlabel={Speedup (\(\times\))}]\ladpanel{mva}{psnr}\betterUR\end{axis}
\begin{axis}[lad, name=c2, at={(c1.south)}, anchor=north, yshift=\dy, xlabel={Speedup (\(\times\))}]\ladpanel{sync}{psnr}\betterUR\end{axis}
\begin{axis}[xmode=log, lad, ladlogx, log ticks with fixed point, name=d2, at={(d1.south)}, anchor=north, yshift=\dy, xlabel={Speedup (\(\times\))}]\ladpanel{mvp}{psnr}\betterUR\end{axis}
\node[anchor=south] at ($(a1.north west)!0.5!(d1.north east)+(0,0.15cm)$) {\pgfplotslegendfromname{ladderlegend}};
\node[anchor=north, font=\footnotesize] at ([yshift=-2pt]a2.outer south -| a2.south) {(a) Hunyuan3D-2.1};
\node[anchor=north, font=\footnotesize] at ([yshift=-2pt]a2.outer south -| b2.south) {(b) MV-Adapter};
\node[anchor=north, font=\footnotesize] at ([yshift=-2pt]a2.outer south -| c2.south) {(c) SyncMVD};
\node[anchor=north, font=\footnotesize] at ([yshift=-2pt]a2.outer south -| d2.south) {(d) MVPainter};
\end{tikzpicture}}
    \caption{Speedup against MV-LPIPS (top) and MV-PSNR (bottom) on (a) Hunyuan3D-2.1, (b) MV-Adapter, (c) SyncMVD, and (d) MVPainter. The dashed line marks an MV-LPIPS of 0.03, at which we compare the speedups. Each curve sweeps one accelerator over its speed parameter, which is \(\lambda\) for \gc{} and the threshold, budget, or interval for the other accelerators, starting from its default setting.}
    \label{fig:speedperf}
\end{figure*}

The correspondence is built once per asset before sampling from the position maps, which give the 3D point seen at each pixel of each view.
Tokens of different views are matched when they show the same surface point, and for each token of \(v\), \(\mathcal{G}_{\mathcal{V}_t\to v}\) returns the weighted average of the transported quantity over its matched tokens in the evaluated views.
Background tokens and tokens without a match in \(\mathcal{V}_t\) receive nothing and keep their previous prediction.
The update of a step is shared between views only in part, so the geometric term is weighted, and we set \(\omega = 1/N\).
An evaluated view is one of the \(N\) views of the object, so the change that it reports contributes one share in \(N\) to the update of another view.
A larger weight replaces more of the content of a view with that of the other views, and we compare the weights in \cref{sec:ablation}.

\section{Evaluation}
\label{sec:exp}

\subsection{Experimental Settings}
\label{sec:setup}
\noindent\textbf{Backbones and data.}
We evaluate on four multi-view texturing backbones, which differ in how the views are denoised together (configurations in the supplement).
Hunyuan3D-2.1 Paint \citep{hunyuan3d21} and MV-Adapter \citep{mvadapter} place the views on the batch axis and connect them through multi-view attention.
SyncMVD \citep{syncmvd} runs a text-to-image model on each view separately and blends the view latents in UV space at every DDPM \citep{ddpm} step.
MVPainter \citep{mvpainter} arranges six views as the tiles of one \(3{\times}2\) grid image.
A view is therefore a batch row or a grid tile, depending on the backbone.
The benchmark eval200 is a frozen list of 200 assets, 100 from GSO \citep{gso} and 100 from Objaverse \citep{objaverse}, and dev20 is a 20-asset subset of it (selection in the supplement).
The study in \cref{sec:preliminary} and the ablations in \cref{sec:ablation} are on dev20, which is held out from all other results on eval200.

\noindent\textbf{Compared methods.}
We compare \gc{} with six training-free accelerators \citep{teacache,magcache,budcache,fora,fastercache,deepcache}. TeaCache, MagCache, BudCache, and FORA are step-level methods, and FasterCache and DeepCache are in-step methods (\cref{sec:preliminary}).
Each of them is swept over its speed parameter, starting from its out-of-the-box configuration, and the supplement tabulates the results of these configurations.

\noindent\textbf{Implementation details.}
Unless stated otherwise, \gc{} runs on the stock sampler of each backbone and selects between 0 and \(N\) views at each step.
The first two steps and the last step are always evaluated on all \(N\) views.
The first two initialize the scores, since \cref{eq:score} needs two evaluations of every view, and the last one evaluates all views once before decoding.
On the 15-step Hunyuan3D-2.1 sampler, three steps are therefore fixed and the selector decides the other twelve.
No periodic refresh is used, since the score of a view that is left out keeps increasing until the view is selected.
We use the transport weight \(\omega=1/N\) and the synchronization weight \(\mu=1\) on all backbones.
The cost table \(c(n)\) is timed once for each backbone on the evaluation machine by timing a step for each \(n\), and nothing is fitted to the model or to data.
The supplement gives the construction of the correspondence and the interface to the sampler.
Every run uses one NVIDIA RTX 4090.

\noindent\textbf{Evaluation metrics.}
MV-LPIPS and MV-PSNR are LPIPS \citep{lpips} and PSNR between the accelerated and the stock renders of each view at the same seed, averaged over views and assets.
SeamErr is the mean absolute color difference between the views of one method at pixels that show the same surface point, and it measures the consistency between views and not the fidelity to the stock model.
Speedup is the wall-clock ratio of the denoising loop, and the supplement gives the details of the metrics.

\begin{figure*}[t]
\centering
\input{Figures/f5_qualitative_tikz}
\caption{Qualitative comparison on Hunyuan3D-2.1, MVPainter, and SyncMVD, with two assets per backbone. For each asset we show one view and a zoom into the red window of that view for the stock model, the compared SOTAs, and \gc{}. The view and the window are the same for the three methods of an asset, and the settings and the speedups of every row are listed in the supplement.}
\label{fig:qual3}
\end{figure*}

\subsection{Main Results}
\label{sec:main}
\noindent\textbf{Quantitative comparison.}
\Cref{fig:speedperf} compares \gc{} with the six accelerators over their sweeps.
The dashed line in the top row marks an MV-LPIPS of 0.03, and we compare the highest speedup that each method reaches below this line.
\gc{} reaches \(2.72\times\) on Hunyuan3D-2.1 and \(4.47\times\) on MV-Adapter, against \(2.21\times\) for BudCache and \(3.65\times\) for DeepCache, which are the best of the compared accelerators on these backbones.
On MVPainter, it reaches \(2.28\times\) against \(1.95\times\) for TeaCache, and on SyncMVD it reaches \(1.40\times\) while no setting of the compared accelerators is below the line.
The advantage grows with the speedup, because the fidelity of \gc{} degrades more slowly.
On Hunyuan3D-2.1, the MV-LPIPS of TeaCache grows from 0.0103 to 0.0961 between \(1.55\times\) and \(2.50\times\), while that of \gc{} grows from 0.0069 to 0.0244 between \(1.51\times\) and \(2.72\times\), and MV-PSNR gives the same ordering.
A step-level accelerator skips all views of a step together, and a higher speedup forces it to skip runs of consecutive steps, during which no view is updated by the denoiser.
\gc{} returns to a view once its predicted error has grown (\cref{eq:score}), so the evaluations that it keeps are spread over the trajectory and over the views, and the other views receive their update by transport.
This difference is largest on MVPainter, whose sampler has 75 steps.
The MV-LPIPS of TeaCache grows from 0.027 at \(1.95\times\) to 0.30 at \(3.89\times\) and its output collapses at higher speedups, while that of \gc{} grows only from 0.029 at \(2.28\times\) to 0.060 at \(13.30\times\).
The size of the advantage follows the design of the backbone.
Hunyuan3D-2.1 uses 15 steps, so one skipped step removes \(1/15\) of the loop and a step-level schedule is coarse.
SyncMVD denoises each view by a separate forward, so a view that is not evaluated does not change the output on the evaluated views, and its ten views give a step more choices than six.
MV-Adapter uses 50 steps, and the curve of \gc{} is below those of the six accelerators over its whole range, with an MV-LPIPS of 0.0070 at \(2.61\times\) against 0.0107 for MagCache at \(2.39\times\) and 0.0100 for DeepCache at \(2.84\times\).
MVPainter uses 75 steps, with which a step-level schedule is fine-grained, and the MV-LPIPS of \gc{} is close to 0.03 already at low speedups (\cref{sec:limits}), so its advantage appears above \(2\times\).

\noindent\textbf{Qualitative results.}
\Cref{fig:qual3} compares the outputs of the stock model, the compared SOTAs, and \gc{} on two assets per backbone.
On Hunyuan3D-2.1, the compared accelerator shifts the color of the texture. It turns the sole of the boot redder and speckled, and it saturates the print on the cereal box.
\gc{} keeps the color of the stock model on both assets.
On MVPainter, the compared accelerator adds blotchy shading to the uniform body of the shark and distorts the collar of the boot, while the output of \gc{} is hard to tell apart from the stock model.
On SyncMVD, the compared accelerator changes the content of the texture. It replaces the figure inside the wheel of the Lego model and turns the mane of the lion yellow-green, while \gc{} keeps the layout and the palette of the stock model.
These examples agree with the curves in \cref{fig:speedperf}.
The compared accelerators apply one approximation to all views of a step, so their error appears as a shift in color or shading over the whole object.
\gc{} spends its evaluations on the views with the largest predicted error and restores the other views from them, so its output stays close to the stock model.
On MV-Adapter, the three outputs are hard to tell apart by eye.

\subsection{Ablation Studies}
\label{sec:ablation}
\begin{table}[t]
    \centering
    \caption{Ablations on Hunyuan3D-2.1 on dev20 at a fixed \(\lambda\). Each row changes one setting of \gc{} and keeps the others at their default values. Random \(\mathcal{V}_t\) draws the evaluated views at random.}
    \label{tab:ablation}
    \footnotesize
    \begin{tblr}{width=\columnwidth, colspec={@{}X[l]cccc@{}}, colsep=3pt, rowsep=0.8pt}
    \toprule
    Setting & Speed\(\uparrow\) & MV-LPIPS\(\downarrow\) & MV-PSNR\(\uparrow\) & SeamErr\(\downarrow\) \\
    \midrule
    \(\omega{=}1/N,\ \mu{=}1\) & 2.11\(\times\) & 0.0248 & 36.63 & 0.0547 \\
    \midrule
    random \(\mathcal{V}_t\) & 2.10\(\times\) & 0.0395 & 33.74 & 0.0589 \\
    \(\mu{=}0\) & 2.13\(\times\) & 0.0289 & 34.47 & 0.0712 \\
    \midrule
    \(\omega{=}0\) & 2.11\(\times\) & 0.0252 & 36.49 & 0.0557 \\
    \(\omega{=}k/N\) & 2.11\(\times\) & 0.0249 & 36.65 & 0.0549 \\
    \(\omega{=}1\) & 2.10\(\times\) & 0.0252 & 36.48 & 0.0542 \\
    shuffled \(\mathcal{G}\) & 2.11\(\times\) & 0.0250 & 36.52 & 0.0554 \\
    \bottomrule
    \end{tblr}
\end{table}

\Cref{tab:ablation} changes one setting of \gc{} at a time on Hunyuan3D-2.1, at a fixed \(\lambda\) that gives a speedup of about \(2.1\times\).
The ablations run on dev20, so their values are not directly comparable with the curves in \cref{fig:speedperf} and \cref{fig:analysis}a, which are on eval200.

\noindent\textbf{Selection.}
A random \(\mathcal{V}_t\) replaces the importance score with a random choice of the same number of views at each step.
It keeps the number of evaluated views and therefore the cost, and it removes only the ranking of the views.
This change has the largest effect on the fidelity.
The MV-LPIPS rises from 0.0248 to 0.0395 and the MV-PSNR falls from 36.63 to 33.74\,dB at the same speedup, so which views are evaluated matters as much as how many.
Setting \(\mu=0\) removes the synchronization term and reduces the selection to \cref{eq:select}.
This change has the largest effect on the consistency between views.
The SeamErr rises from 0.0547 to 0.0712, and the MV-LPIPS rises from 0.0248 to 0.0289.
The constraint therefore improves both the consistency between views, which it targets, and the fidelity.

\noindent\textbf{Restoration.}
The transport restores the views that are not evaluated and does not change which views are evaluated, so the speedup is the same in these rows and they differ only in the quality of the restoration.
With \(\omega=0\), a view keeps its previous prediction and receives nothing from the evaluated views.
The coverage weight \(\omega=k/N\), for a token that is seen by \(k\) evaluated views, and the full strength \(\omega=1\) give the geometric term more weight than the default.
The shuffled \(\mathcal{G}\) is the control for the geometry. It exchanges the matched tokens among the foreground tokens of each view, which keeps the sparsity, the weights, and the cost and removes only the geometry.
The MV-LPIPS of these rows is between 0.0248 for the default and 0.0252, and their MV-PSNR and SeamErr are within 0.17\,dB and 0.0015 of each other.
The way a view is restored therefore has a small effect, and the gain of \gc{} comes from the selection.

\subsection{Analysis}
\label{sec:analysis}
\begin{figure}[t]
\centering
\input{Figures/f4_ladder_data}%
\resizebox{\columnwidth}{!}{\begin{tikzpicture}
\definecolor{gcred}{RGB}{200,50,45}
\definecolor{c1}{RGB}{54,125,189}\definecolor{c2}{RGB}{77,160,95}\definecolor{c3}{RGB}{140,100,180}
\definecolor{c4}{RGB}{214,120,40}\definecolor{c5}{RGB}{120,120,120}\definecolor{c6}{RGB}{170,140,60}
\definecolor{orc}{RGB}{54,125,189}\definecolor{ffc}{RGB}{214,120,40}
\pgfplotsset{ana/.style={scale only axis, width=2.85cm, height=2.75cm, grid=major, grid style={black!10},
  tick label style={font=\scriptsize}, label style={font=\footnotesize}, xlabel={Speedup (\(\times\))}, xlabel shift=-3pt, ylabel shift=-4pt}}
\newcommand{\anabetter}{\draw[-{Stealth[length=3.5pt]},black!55,line width=0.6pt] (axis description cs:0.70,0.90) -- (axis description cs:0.90,0.70);
  \node[font=\scriptsize,text=black!55,anchor=south,inner sep=1.5pt] at (axis description cs:0.76,0.90) {better};}
\begin{axis}[ana, name=axa, enlarge x limits=0.06, enlarge y limits=0.1, scaled y ticks=false,
  ytick={0.03,0.035,0.04,0.045}, yticklabel style={/pgf/number format/fixed, /pgf/number format/precision=3},
  ylabel={SeamErr \(\downarrow\)},
  legend to name=analegA, legend columns=4,
  legend style={font=\scriptsize, draw=none, /tikz/every even column/.append style={column sep=4pt}, inner sep=0pt},
  legend image code/.code={\draw[#1,mark repeat=2,mark phase=2] plot coordinates {(0cm,0cm) (0.17cm,0cm) (0.34cm,0cm)};}]
\addplot[c1, mark=o, mark size=1.5pt, line width=0.7pt] table[x=speed,y=seam]{\ladhyTea};
\addplot[c2, mark=square, mark size=1.4pt, line width=0.7pt] table[x=speed,y=seam]{\ladhyMag};
\addplot[c3, mark=triangle, mark size=1.8pt, line width=0.7pt] table[x=speed,y=seam]{\ladhyBud};
\addplot[c4, mark=diamond, mark size=1.8pt, line width=0.7pt] table[x=speed,y=seam]{\ladhyFORA};
\addplot[c5, mark=pentagon, mark size=1.7pt, line width=0.7pt] table[x=speed,y=seam]{\ladhyFaster};
\addplot[c6, mark=x, mark size=1.9pt, line width=0.7pt] table[x=speed,y=seam]{\ladhyDeep};
\addplot[gcred, mark=star, mark size=2.6pt, line width=1.3pt] table[x=speed,y=seam]{\ladhySeamGC};
\legend{TeaCache,MagCache,BudCache,FORA,FasterCache,DeepCache,\textbf{ViewCache}}
\anabetter
\end{axis}
\begin{axis}[ana, name=axb, at={(axa.east)}, anchor=west, xshift=1.30cm, xmin=1.85, xmax=2.45, ymode=log, ymin=0.009, ymax=0.2,
  xtick={1.9,2.1,2.3}, ylabel={MV-LPIPS \(\downarrow\)},
  ytick={0.01,0.02,0.05,0.1}, yticklabels={0.01,0.02,0.05,0.1}, clip=false]
\newcommand{\garrow}[5]{%
  \addplot[#1, line width=1.0pt, -{Stealth[length=3.5pt]}, shorten >=2pt, forget plot] coordinates {(#2,#3) (#4,#5)};
  \addplot[#1, only marks, mark=o, mark size=1.9pt, line width=0.7pt, forget plot] coordinates {(#2,#3)};
  \addplot[#1, only marks, mark=*, mark size=2.1pt, forget plot] coordinates {(#4,#5)};}
\garrow{orc}{2.00}{0.0261}{2.30}{0.0156}
\garrow{orc}{2.21}{0.0487}{2.27}{0.0152}
\garrow{ffc}{1.98}{0.1289}{2.19}{0.0143}
\node[font=\scriptsize, text=orc, anchor=north, inner sep=3.5pt, xshift=-5pt] at (axis cs:2.00,0.0261) {TeaCache};
\node[font=\scriptsize, text=orc, anchor=south, inner sep=2.5pt] at (axis cs:2.21,0.0487) {MagCache};
\node[font=\scriptsize, text=ffc, anchor=west, inner sep=3pt] at (axis cs:1.98,0.1289) {FORA};
\anabetter
\end{axis}
\node[anchor=south,inner sep=0pt] at ($(axa.north west)!0.5!(axb.north east)+(-0.25cm,0.14cm)$) {\pgfplotslegendfromname{analegA}};
\node[anchor=north, font=\footnotesize] at ([yshift=-2pt]axa.outer south -| axa.south) {(a) Cross-view consistency};
\node[anchor=north, font=\footnotesize] at ([yshift=-2pt]axa.outer south -| axb.south) {(b) Combination};
\end{tikzpicture}}
\vspace{-0.5cm}
\caption{Analysis on Hunyuan3D-2.1. (a) SeamErr against speedup. (b) Each arrow goes from a step-level accelerator alone (hollow) to the same accelerator with \gc{} added (filled).}
\label{fig:analysis}
\vspace{-0.25cm}
\end{figure}

\noindent\textbf{Cross-view consistency.}
\Cref{fig:analysis}a reports SeamErr on Hunyuan3D-2.1 for the sweeps of \cref{fig:speedperf}.
The SeamErr of \gc{} stays between 0.031 and 0.035 and does not grow with the speedup, while that of TeaCache, MagCache, BudCache, and FORA grows as more steps are skipped, up to 0.046 for FORA.
The synchronization constraint keeps this consistency, since it penalizes overlapping views that are left with different errors.
DeepCache is the closest, between 0.033 and 0.035, at a higher MV-LPIPS.
FasterCache has the lowest value, 0.032 at \(1.94\times\), at an MV-LPIPS of 0.1058, so SeamErr alone does not indicate quality.

\noindent\textbf{Combination with step-level accelerators.}
\label{sec:generality}
We add \gc{} to three step-level accelerators on Hunyuan3D-2.1, namely TeaCache, MagCache, and FORA (\cref{fig:analysis}b).
The step-level accelerator decides which steps run, \gc{} decides which views run on those steps, and the decision rule of the accelerator is unchanged.
The rules and settings of \gc{} are the same for every accelerator, namely the importance score of \cref{eq:score}, the transport weight \(\omega=1/N\), and the synchronization weight \(\mu=1\), and none of them is tuned per accelerator.
The cost table \(c(n)\) is a measurement and not a tuned parameter. It is timed once for each backbone and each accelerator, since an accelerator changes the time of a step, and the value of \(\lambda\) can therefore differ across accelerators.
With \gc{} added, each of the three accelerators is both faster and closer to the stock model.
TeaCache goes from \(2.00\times\) to \(2.30\times\) while its MV-LPIPS drops from 0.0261 to 0.0156, MagCache goes from \(2.21\times\) to \(2.27\times\) with 0.0487 to 0.0152, and FORA goes from \(1.98\times\) to \(2.19\times\) with 0.1289 to 0.0143.

\section{Limitations}
\label{sec:limits}

\gc{} depends on position maps, or an equivalent geometric representation, to relate the views, so surface regions seen by few evaluated views receive fewer transported updates and rely on the full-view steps.
On MVPainter, its MV-LPIPS rises to about 0.03 as soon as views are left out, so FasterCache and TeaCache are closer to the stock model below \(2\times\), and we leave the analysis of this floor to future work.
\gc{} reduces the denoiser computation only, so the end-to-end gain follows the share of the denoising loop in the pipeline time, which the supplement reports.

\section{Conclusion}
We presented \gc{}, a training-free accelerator for multi-view texture diffusion that works at the view level.
It builds on two observations, namely that the views of a step differ in how much they change and that they are similar on the surface that they share.
\gc{} selects any number of views to evaluate at each step by an importance score under a synchronization constraint, and it restores the other views without the network.
On four backbones, it achieves a better trade-off between speedup and fidelity than six training-free accelerators, and it can be combined with step-level ones.

\section*{LLM Usage Statement}

We used Claude (Anthropic) to assist with writing and editing text (language, restructuring), refining related work, improving visualization code. The authors reviewed and verified the generated content and remain responsible for the technical correctness and scientific claims of this work.

{
    \small
    \bibliographystyle{ieeenat_fullname}
    \bibliography{main}
}

\end{document}